\documentclass[twoside]{article}

\usepackage[accepted]{aistats2023}
\usepackage{amssymb}
\usepackage{graphicx}
\usepackage{caption}
\usepackage{subcaption}
\usepackage{stmaryrd}
\usepackage{hyperref}

\usepackage[round]{natbib}

\setcitestyle{authoryear}

\usepackage{amsthm}

\newtheorem{theorem}{Theorem}[section]
\newtheorem{proposition}{Proposition}[section]

\newtheorem*{remark}{Remark}

\newcommand{\N}{\mathbb{N}}
\newcommand{\R}{\mathbb{R}}

\newcommand{\E}{\mathbb{E}}

\begin{document}

%

%

\twocolumn[

\aistatstitle{Singular Value Representation: A New Graph Perspective On Neural Networks}

\aistatsauthor{ Dan Meller \And Nicolas Berkouk }

\aistatsaddress{ École Polytechnique \\ EPFL - Topology and Neuroscience Lab \And  EPFL - Topology and Neuroscience Lab } ]

\begin{abstract}
  We introduce the Singular Value Representation (SVR), a new method to represent the internal state of neural networks using SVD factorization of the weights. This construction yields a new weighted graph connecting what we call spectral neurons, that correspond to specific activation patterns of classical neurons. We derive a precise statistical framework to discriminate meaningful connections between spectral neurons for fully connected and convolutional layers. 
  To demonstrate the usefulness of our approach for machine learning research, we highlight two discoveries we made using the SVR. First, we highlight the emergence of a dominant connection in VGG networks that spans multiple deep layers. Second, we witness, without relying on any input data, that batch normalization can induce significant connections between near-kernels of deep layers, leading to a remarkable spontaneous sparsification phenomenon. 
\end{abstract}

\paragraph{Code:} A Python implementation of the SVR can be found at \url{https://github.com/danmlr/svr}

\section{INTRODUCTION}

\subsection{Motivation}
Following the motivation for the first perceptron by \citet{perceptron}, artificial neural networks are often introduced as a stylized version of their biological counterpart. This has led many to represent neural networks by the graph of neurons connected by edges whose weights correspond to tensor entries. This famous representation provides an overview of calculations performed by the network and clearly highlights the role of activation functions. However, this representation fails to identify collective patterns in neural activations. We instead argue that the focus should be shifted to the successive linear maps which define the network. We propose to use Singular Value Decomposition (SVD) of those maps to identify meaningful input and output directions corresponding to collective activation patterns of neurons. We further investigate the interaction of those directions across deep layers, which yields a new graph representation of neural networks. This graph provides a high-level overview of the network that allows to witness the emergence of global phenomenons across deep layers as highlighted in the last section.




\subsection{Related Work}

Visualizing the internal computational pipeline of neural networks has been widely studied. However, most approaches rely on studying the activations induced by input data as it flows through the network \citep{revacnn,activis,flows}. Moreover, those approaches mainly focus on creating user-friendly interfaces that are not suited for theoretical analysis.  
On the other hand, SVD factorization of weight tensors has also been widely explored for its potential to compress neural networks while still maintaining good accuracy levels \citep{svd_acoustic,svd_conv,compression_survey}. More specifically, the factorization technique we use for convolutional layer has already been described in \citet{svd_conv,Jaderberg,Idelbayev,Wen2017,global_svd} but only for compression purposes. The originality of our approach lies in the use of this compression technique to identify local structure \emph{accross} layers.


\section{THE SVR FOR FULLY CONNECTED LAYERS}

\paragraph{Definition of SVD:} \textit{See \citet{svd_history} for a historical overview}. 
The singular value decomposition (SVD) allows factoring any real matrix, in the form ${A=USV^T}$ where $U$ and $V$ are matrices whose columns are orthonormal vectors\footnote{But not necessarily a base. If the linear map has lower output dimension than input dimension, $V$ will not be a square matrix.} and where $S$ is a diagonal matrix with strictly positive entries in decreasing order that are called singular values. 

\subsection{Intuitive Explanation} 

The SVD of a matrix can be interpreted as a computational pipeline. It breaks down into 3 parts the action of a linear map: in a first stage, orthogonally independent features are computed by dotting the input with the columns of $V$. Those features are then scaled according to their singular values ($S$). The output is finally obtained by an orthogonal base change\footnote{This part may seem obvious to the reader, however this point of view is key to understand the generalization to convolutional layers in the next sections}.
The $k$-th column of $V$ and the $k$-th column of $U$ can be seen respectively as  the input and output representation of the same underlying object which we call the $k$-th \textbf{spectral neuron}\footnote{The authors are aware that the term spectral is slightly misused here but we found it to be less confusing than "singular neurons".}. Each spectral neuron thus has both an \textbf{input and an output representation} which correspond to collective activation patterns of classical neurons in respectively the input and output layer. The idea behind the SVR is that the relation between spectral neurons in consecutive layers captures key insights about deep information processing.

\subsection{Formal Definition}

A simple fully connected and bias-free neural network can be seen as a sequence of linear maps $(A_i)_{0 \leq i \leq n}$ intertwined with a non-linear activation function $\psi$ applied to every vector component (ReLU, Sigmoid, hyperbolic tangent etc.). The network corresponds to the composition: $\left( f \circ A_n \circ \psi \circ A_{n-1} \circ ... \circ \psi \circ A_0 \right) $ where $f$ is the final function applied to obtain the output (often argmax for a classification task). 
The SVR of a fully connected neural network can be formally defined as a graph:
\paragraph{Vertices (Spectral Neurons)}
 Every linear map can be factored using SVD: $A_i=U_i S_i V_i^T$. The $k$-th spectral neuron in layer $i$ is defined by the triplet: $$\left((U_i)_{\_,k} , (S_i)_k, (V_i)_{\_,k} \right)$$ where $(U_i)_{\_,k}$ (resp. $(V_i)_{\_,k}$) is the $k$-th column of $U_i$ (resp $V_i$) and where $(S_i)_k \in \R$ is the $k$-th singular value of $A_i$.

\paragraph{Edges (Connecting Spectral Neurons)}
The matrix $|V_{i+1}^T U_i|^2$, whose entries are the square of the entries of the matrix $V_{i+1}^T U_i$, can be seen as a weighted adjacency matrix between the spectral neurons of layer $i$ and those of layer $i+1$. Each coefficient in this matrix correspond to an edge weight which indicates the interaction intensity between two spectral neurons respectively in layer $i$ and layer $i+1$.
\\ 
\\ 
This structure only captures the information flow between spectral neurons and cannot be used for calculations, as the non-linearity $\psi$ is omitted in this representation\footnote{If we let $n$ denote the output dimension of $U_i$, the adjacency matrix would not be changed if $\psi$ was replaced by $Q \circ \psi \circ Q^T$ where $Q \in O_n(\R)$.}. This method may seem overly simple to describe neural networks, as it amounts to performing a locally linear approximation between consecutive layer by removing the non-linearity. It is therefore quite surprising that this method allows to witness the emergence of meaningful global structures across layers, as we shall see in the next sections. 

\begin{figure}
\centering
\includegraphics[scale=0.22]{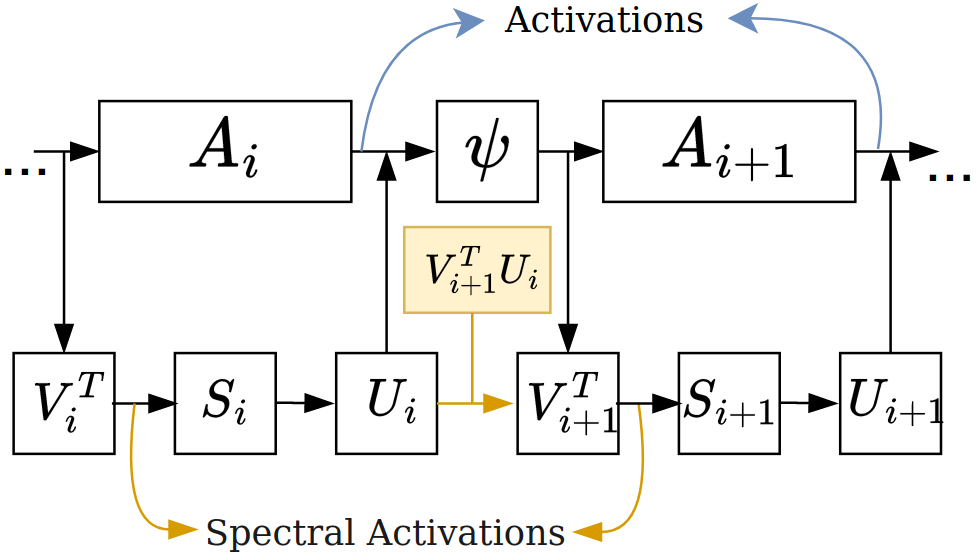}
\caption{SVR Diagram Overview}
\label{explanationSVR}
\end{figure}

\paragraph{Activations and Spectral Activations:}
The activation of neurons ($(X_i)_{1 \leq i \leq n+1}$) induced by an input $X_0$ is the set of vectors obtained after each linear map $X_{i+1} := \left( A_{i} \circ \psi \circ A_{i-1} ... \circ A_0 \right) (X_0)  $. It is also possible to define activation over spectral neurons by applying the orthonormal projection induced by the matrices $(V_i)_{0 \leq i \leq n}$. The \textbf{spectral activations} $(Y_i)_{0 \leq i \leq n}$ are thus given by: $Y_i := \left( V_i^T \circ \psi \right)(X_i) $. This is summarized in Figure \ref{explanationSVR}. 

\subsection{Elementary Properties} 

\paragraph{Upper Bound on Coefficients} Each coefficient in the adjacency matrix $|V_{i+1}^T U_i|^2$ is the square of the scalar product between two normal vectors : a column of $V_{i+1}$ and a column of $U_{i}$. Each coefficient is thus bounded by 1 thanks to the Cauchy-Schwarz inequality. 

\paragraph{Positive Scale Invariance of Adjacency Matrices} The adjacency coefficients in the SVR are invariant under scaling of the weights of the neural network by a positive scalar. That is, for all sequence $(\lambda_i)_i$, with $\lambda_i >0$, the fully connected networks associated to the collections of linear maps $(A_i)_i$ and $(\lambda_i \cdot A_i)_i$ have the same SVR adjacency matrices. This is simply because singular values are not taken into account. This is especially remarkable because most classification networks using ReLU functions also exhibit this property \citep{neuralTeleportation}: scaling their weights by a positive scalar will not affect their behavior. This is due to the fact that both the linear maps of the network and the ReLU function are equivariant under multiplication by a positive scalar whereas the last layer is often invariant to such a scaling (argmax for example). 

\subsection{Statistical Significance Threshold} 

The adjacency matrix between two consecutive linear maps $A_i$ and $A_{i+1}$ is given by $|V_{i+1}^T U_i|^2$. It is worth asking which coefficients in this matrix are significant and which are simply induced by noise. Since each coefficient in this matrix is the square of the scalar product between two unit vectors, a simple probabilistic framework can be used to obtain a threshold above which coefficients are considered significant within a certain confidence level. The essential question is, if we draw randomly two independent vectors $X,Y \sim \mathcal{U}(\mathbb{S}^{n-1})$ according to the uniform distribution on the sphere in $\R^n$, what is the distribution of the square of their scalar product $\langle X,Y \rangle^2 $ ?  The expected value can be computed by noticing that rotational invariance implies that the distribution of $\langle X,Y \rangle$ is the same than that of $\langle X,e_1 \rangle =x_1$ where $e_1$ is the first vector of the canonical base. Exploiting permutation invariance between the coordinates yields the result: \footnote{Using similar symmetry techniques, it is possible to compute all moments of $\langle X, Y \rangle$. For the fourth moment, we get (see Proposition \ref{prop:mu4}): 
 $$\E(\langle X, Y \rangle^4)=\frac{3}{n^2+2n} \underset{n \rightarrow +\infty}{\simeq} 3\ \E(\langle X, Y \rangle^2)^2$$}  
\begin{align*}
 \E(\langle X,Y \rangle^2) = \E(x_1^2) &= \frac{1}{n}\sum_{i=1}^n \E(x_i^2) \\ 
 &= \frac{1}{n}\E\left(\sum_{i=1}^n x_i^2\right) = \frac{1}{n}. \
\end{align*}

Using this scaling, we can gain a deeper understanding of the distribution of $\langle X,Y \rangle^2$. The following result yields in the limit $n\longrightarrow + \infty$ a rescaled  $\chi_2(1)$ distribution (which is simply the distribution of a squared normal random variable). 
\begin{theorem}
\label{chi2_1} \citep{BorelIntroductionG,Finetti} If $X\sim\mathcal{U}(\mathbb{S}^{n-1})$ and $Y \in \mathbb{S}^{n-1}$ then we have the convergence in distribution:
$$ n \langle X,Y\rangle^2 \underset{n \longrightarrow +\infty}{\Longrightarrow} \chi_2(1).$$ 
\end{theorem}

\begin{proof}
See Appendix, section \ref{proof_chi2_1}. 
\end{proof}

Building on this analysis, we will now assume as a null hypothesis that $\langle X,Y\rangle^2 \sim \frac{1}{n} \chi_2(1)$. 

If we now let $n$ be the dimension of the columns of $U_i$ (and $V_{i+1}$) and denote by $Q_n(p)$ the quantile function of $\frac{1}{n} \chi_2(1)$, we can consider that all coefficients of $|V_{i+1}^T U_i|^2$ with a value above $Q_n(p)$ are significant with a probability $p$ confidence. We provide computation of higher moments of $\langle X,Y \rangle$ in Appendix \ref{A:momentsFC}. 

\section{BLOCK MATRIX STRUCTURE}

\subsection{Overview}

We use a simple setup to illustrate what has been described above. 
We train for 5 epochs a fully connected neural network with ReLU activations to classify the MNIST dataset (reaching 94\% test accuracy) \citep{mnist}. We use the architecture: $[28\times28,40,40,40,10]$ (number of neurons in each layer) with no bias.

\captionsetup[figure]{justification=centering}
\begin{figure*}
\centering
\captionsetup[subfigure]{justification=centering}
\begin{subfigure}{.5\textwidth}
  \centering
  \includegraphics[width=0.964\linewidth]{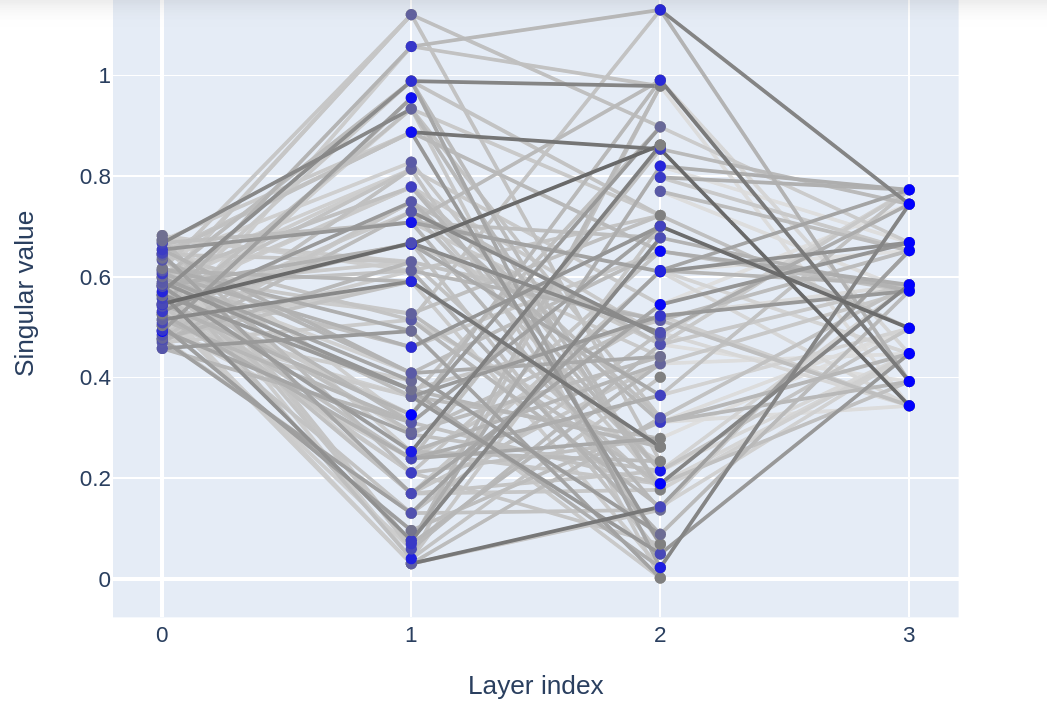}
  \caption{Before Training - Randomly Initialized Network}
  \label{InitSVR}
\end{subfigure}%
\begin{subfigure}{.5\textwidth}
  \centering
  \includegraphics[width=1\linewidth]{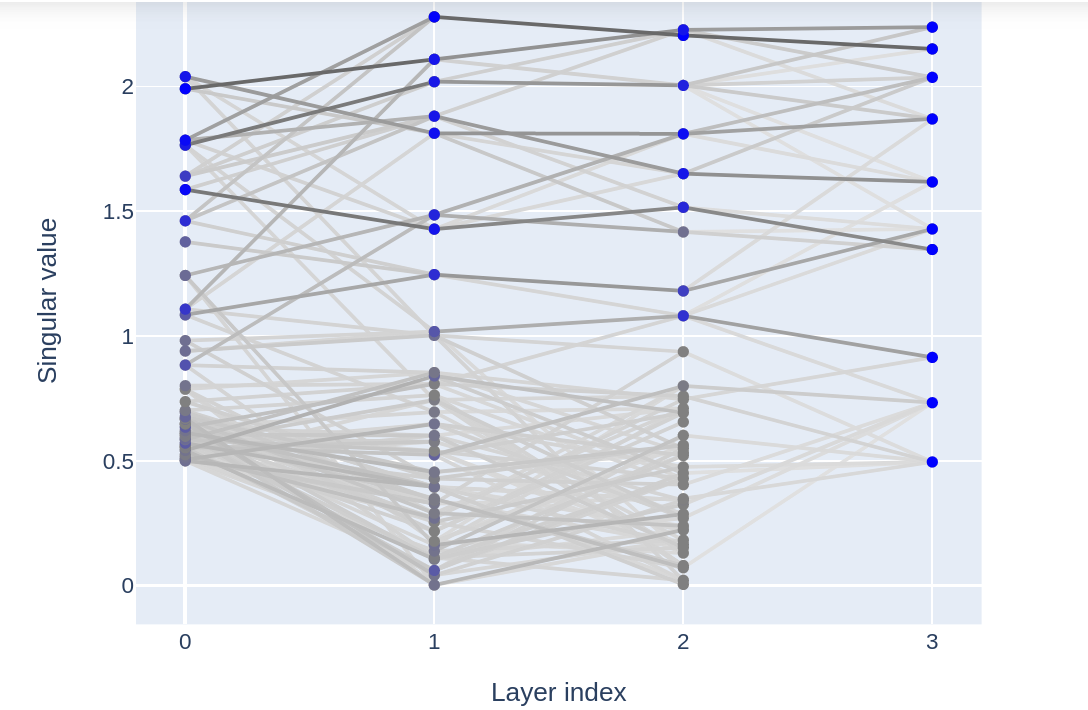}
  \caption{After Training (5 Epochs on MNIST)}
  \label{EndSVR}
\end{subfigure}

\caption[Caption for LOF]{SVR of Neural Network Before (a) and After (b) training - architecture: $[28\times28,40,40,40,10]$ \protect\footnotemark}
\label{vgg-resnet}. Legend: gray-scale of edges correspond to intensity of entries of $|V_{i+1}^T U_i|^2$; spectral neuron color intensity is explained at the end of section \ref{sec:internal_dim} 
\end{figure*}

\footnotetext{The color of each vertex measures the deviation of the associated column in the next adjacency matrix to a random distribution. More precisely, it corresponds to the maximum deviation of the cumulative sum of scalar products to the cumulative sum of $1/n$. The same idea is used to describe internal dimensions below.}

Figure \ref{InitSVR} and \ref{EndSVR} depict respectively the SVR representation of the network before and after training. The $y$-axis corresponds to singular values and the $x$-axis to the layers. Only edges which are considered significant are depicted, and their color indicates the magnitude of their weight. We use a probability threshold $p=0.15$ which means that in a random setting, only $15\%$ of edges would be shown. Remember that each layer on this graph represents a linear map of the network, we thus have 4 layers corresponding to the maps: $\R^{28\times 28} \longrightarrow \R^{40}, \R^{40} \longrightarrow \R^{40}, \R^{40} \longrightarrow \R^{40}$ and $\R^{40} \longrightarrow \R^{10}$. The maximal rank of each of those maps is $40$ except for the last one which is at most of rank $10$. Since each vertex represents a singular value in the SVD of a linear map, we have $40$ vertices in the layers 0,1,2 and $10$ vertices in layer 3.

\begin{figure}
    \centering
    \includegraphics[scale=0.25]{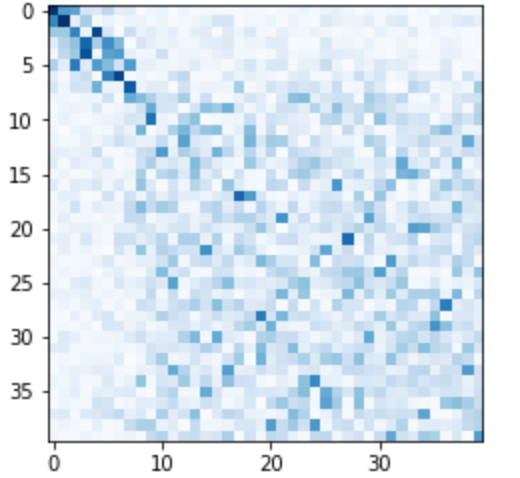}
    \caption{Adjacency Matrix Between Layer 1 and 2 in the Previous SVR Graph - Architecture: $[28\times28,40,40,40,10]$}
    \label{adjacency}
\end{figure}

We can notice that the SVR of a trained network yields a graph with a very specific structure compared to a random baseline. More specifically, we observe the emergence of highly significant connections between spectral neurons of high singular values. The bottom part of the graph appears to be randomly connected with less significant connections. This double structure in the graph becomes even clearer if one has a look at the adjacency matrices that connect consecutive layers (Figure \ref{adjacency}). We can observe a $4$ block structure which sharply separates two types of spectral neurons. If we prune the spectral neurons in the bottom part (precisely defined in section \ref{sec:internal_dim}), which amounts to performing an SVD compression of the layers, we do not observe any changes in the network's test accuracy beyond statistical fluctuations of the order of 1\%. This suggests that the network only uses a small fraction of all the dimensions available, which is the key idea behind \textit{low-rank matrix factorization} \citep{svd_conv,compression_survey,svd_compression}. The top-structure of the SVR graph could also be viewed as an internal trained subnetwork which is very reminiscent of the lottery ticket hypothesis \citep{lottery}.

\vfill

\subsection{Internal Dimensions}
\label{sec:internal_dim}
\begin{figure}

\begin{subfigure}{.25\textwidth}
  \centering
  \includegraphics[width=0.85\linewidth]{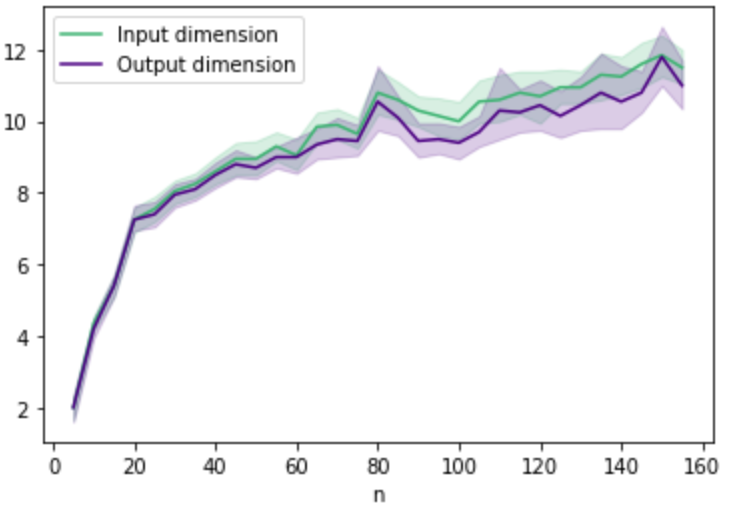}
  \caption{First Adjacency Matrix}
  \label{fig:sub1}
\end{subfigure}%
\begin{subfigure}{.25\textwidth}
  \centering
  \includegraphics[width=0.85\linewidth]{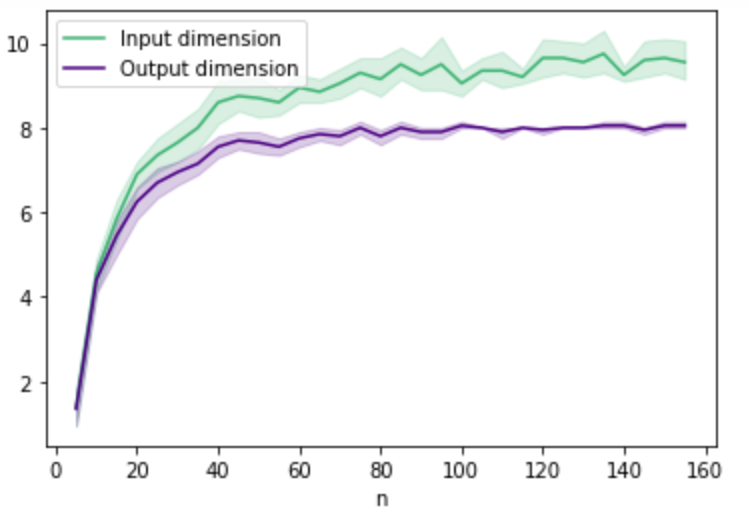}
  \caption{Second Adjacency Matrix}
  \label{fig:sub2}
\end{subfigure}
\caption{Internal Dimensions Used as Functions of $n$ (Averaged over 20 Runs) }
\label{internal_dims}
\end{figure}

One may wonder if the dimensions of the top-left block in Figure \ref{adjacency} change if the number of neurons (\textit{i.e.} the number of available dimension) changes. Does the network use more space if we give it more space ? \\ 
We first have to precisely define how to compute the dimension of the top-left block. Let us consider the adjacency matrix $A$ associated with two consecutive linear maps in a network $f : \R^m \longrightarrow \R^n$ and $g : \R^n \longrightarrow \R^q$. Let us call $B$ the matrix with the same shape as $A$ and for which 

\begin{equation}
    B_{i,j} = \sum_{k\leq i}\sum_{ l \leq j} A_{k,l}.
\end{equation}

One can provide a geometrical interpretation for coefficients of $B$. 
If $f$ has left singular vectors $U_1,...,U_m$ and $g$ has right singular vectors $V_1,...,V_q$ the adjacency coefficient is given by $A_{i,j}=\langle V_i, U_j \rangle^2$. 
If we call $F_j := \text{Span}(U_1,...,U_j)$ and $G_i := \text{Span}(V_1,...,V_i)$, then $B_{i,j}$ corresponds to the square of the Frobenius norm of the composition of orthogonal projections on those subspaces\footnote{Given $E$ a vector subspace of $\R^d$, we let $p_E : \R^d \to \R^d$ be the orthogonal projection onto $E$.}.

\begin{proposition}
The following equality holds: \begin{equation}\label{eq:B}
    {B_{i,j}=\left|\left| p_{F_j} \circ p_{G_i} \right|\right|_F^2 = \left|\left| p_{G_i} \circ p_{F_j} \right|\right|_F^2 }.
\end{equation}
\end{proposition}
\begin{proof}
This result follows from a direct application of definitions. The composition in equation \ref{eq:B}  can be written in any order because orthogonal projections are symmetric and transposition does not change the Frobenius norm. 
\end{proof}

Intuitively, $B_{i,j}$ captures how much $F_j$ matches with $G_i$.  

Recall from the previous section that under a random baseline hypothesis, the expected value of each coefficient in the adjacency matrix is $1/n$. Hence, under the null hypothesis\footnote{Under which $U$ and $V$ are independent and uniformly sampled on $O_{\min(m,n),n}(\R) \times O_{n,\min(n,q)}(\R)$}, we expect $B_{i,j} \simeq \frac{1}{n}ij$. We will thus define the internal dimensions of the adjacency layer to be: 
$$ d_{\text{out}},d_{\text{in}} := \underset{i,j}{\text{argmax}} \left( B_{i,j}-\frac{1}{n}ij \right).$$
The internal dimensions of the matrix are those that maximize the deviation of the cumulative sum of scalar products from its random baseline. We provide in Appendix \ref{sec:internal_dim_validation} an empirical validation of this definition in a controlled setting. 

 In order to identify quickly internal dimension for each layer in a visual way, we set the color intensity of spectral neuron $j$ to be\footnote{Color intensities are then normalized for each layer by their maximums in order to have numbers between 0 and 1} :
 $$ \text{color\_intensity}_j = \underset{i}{\text{max}} \left( \sum_{k \leq i}A_{k,j}-\frac{1}{n}i \right) $$
 Taking the argmax instead of the max would yield the output dimension associated to the matrix $ \begin{pmatrix} A_{1,j} \\ ... \\A_{q,j} \end{pmatrix} $. 
 This can be visualized in Figure \ref{EndSVR} where the bluest spectral neuron in layer 1 correspond to the horizontal dimension of the upper left block in Figure \ref{adjacency}. This approach allows to clearly witness sharp transitions that correspond to internal dimensions while the use of a continuous criterion \footnote{instead of simply using two colors to differentiate spectral neurons that are below or above the output internal dimensions} still provides meaningful insights when no block structure can be identified in the adjacency matrix. 

\paragraph{Experiment}
We consider a set of fully connected networks that have an architecture of the form $[28 \times 28, n,n,10]$. Each network will thus have two adjacency matrices: one that connects the maps $\R^{28\times28} \longrightarrow \R^n$ to $\R^n \longrightarrow \R^n$ and one that connects the maps $\R^n \longrightarrow \R^n$ to $\R^n \longrightarrow \R^{10}$. 
We train each network for 5 epochs to classify the \textit{FashionMNIST} \citep{fashionMNIST} dataset. We obtain test accuracies ranging from $81\%$ (for $n=5$) to $87\%$ we do not observe any significant increase of the test accuracy for $n\geq 40$. Figure \ref{internal_dims} depicts how internal dimensions change as $n$ varies.

 Overall, it seems that the number of dimensions used by the network \textbf{does not depend} on the number of available dimensions \textbf{as long as the latter is sufficiently high}. Even if we see a slight increasing tendency for the first adjacency matrix, the variation of internal dimensions is extremely small compared to the variations of $n$. 
 Interestingly, the best accuracies are reached when the network has more dimensions than what it will end up using. This extra capacity may be crucial for an efficient exploration of the optimization space during training. 
We can also observe that we almost always have  ${d_{\text{out}} \leq d_{\text{in}}}$ for both matrices. This is consistent with the idea that as one moves towards the later layers, the network compresses the data into a lower dimensional subspace.

\section{THE SVR FOR CONVOLUTIONAL LAYERS}

Even though most modern neural network architectures include fully-connected heads, they are almost always used in combination with other types of layers. This motivates the following generalization to convolutional layers, which are almost ubiquitous in artificial vision networks.

\subsection{Intuitive Explanation}

A fully connected layer between an input space of dimension $i$ and an output space of dimension $o$ can be seen as a collection of $o$ applications from $\R^i$ to $\R$. 
Similarly, if we denote by $F$ the vector space of $2D$ images of a fixed shape, a convolutional layer between an input space with $i$ channels and an output space with $o$ channels can be seen as a collection of $o$ applications from $F^i$ to $F$\footnote{We consider that convolution does not change the size of the image (using 0-padding) on borders.}. The analogy between fully connected and convolutional layers is summarized in Appendix (Table \ref{tab:fc-cnn}). 
\\ 
The factorization we introduce in the next section can again be interpreted as a computational pipeline that breaks down the convolution operation into three steps, albeit at the price of a slightly more convoluted mechanism. Starting from the input $x \in F^i$,  $M$ spectral activation images $y \in F^M$ are computed by performing a 2D-convolution of the input $x$ with all $M$ filters corresponding to the input representation of the spectral neurons ($V$). The spectral activations $y \in F^M$ are then scaled according to the singular values contained in $S$. The output for a specific channel $k \in \llbracket 1,o \rrbracket$ is finally computed by performing a linear combination of scaled spectral images using coefficients from $U$: $\sum_{m =1}^M u_{k,m}s_m y_m \in F$.

\subsection{Formal Definition}
The weights of a 2D-convolutional layer are given in the form of a tensor $T\in \R^{o \times i \times K \times K}$ where $K$ is the kernel size (typically 3, 5 or 7).

\paragraph{Vertices (Spectral Neurons)} To compute the SVR we first flatten down the tensor to 2 dimensions $\bar{T} \in \R^{o \times iK^2}$. We then compute the classical SVD: $US\bar{V}^T=\bar{T}$. Any singular value in $S$ corresponds to a spectral neuron of the convolutional layer. Let $M \in \N$ be the number of such values for layer $T$. The input representations associated to spectral neurons, are obtained by unflattening $\bar{V} \in \R^{iK^2 \times M}$ to $V\in \R^{i \times K \times K \times M}$ which should be seen as $M$ convolutional filter in $\R^{i \times K \times K}$. The output representations are simply given by the columns of $U$.

\paragraph{Edges (Connecting Spectral Neurons)}
To understand how to define correctly the adjacency structure, one has to understand how the activation of the spectral neuron $n$ in layer $q+1$ (before scaling) $z_n \in F$ is computed from spectral activations $(y_m) \in F^M $ in layer $q$ (after scaling) if the network did not have any non-linearity. We have:\footnote{Here $V\in \R^{o \times K \times K \times M'}$ is obtained from the SVD of layer $q+1$ while $U\in \R^{o \times M}$ comes from the SVD of layer $q$. Since $V$ is a 4D-tensor, $V_{k,n}$ for $k \in \llbracket 1,o \rrbracket$ corresponds to a filter in $\R^{K \times K}$ while $u_{k,m}$ for $m \in \llbracket 1,M \rrbracket$ is simply a scalar since $U$ is just a matrix. The notation $\star$ corresponds to 2D-convolution. }

\begin{align*}
z_n&= \sum_{k=1}^o  V_{k,n} \star\left( \sum_{m=1}^M u_{k,m}  y_m\right) \\ 
&= \sum_{m=1}^M \left( \sum_{k=1}^o u_{k,m}V_{k,n} \right) \star y_m.
\end{align*} 

This simple calculation shows that, thanks to the linearity of the convolution operation, $z_n$ is simply the sum of spectral activations in the previous layer $(y_m)$ convoluted by a $K \times K $ effective filter ${A_{n,m} = \sum_{k=1}^o u_{k,m}V_{k,n} \in \R^{K \times K}}$. We thus define the adjacency coefficient between spectral neurons $m$ and $n$ to be: 
\begin{align*}
 a_{n,m} :&= \left|\left|A_{n,m}\right|\right|_F^2 \\ 
 & =\left|\left|\sum_{k=1}^o u_{k,m}V_{k,n}\right|\right|_F^2 \\ 
 &=\sum\limits_{\substack{z \in K \times K}} \left( \sum_{k=1}^o u_{k,m} v_{k,z,n} \right)^2. 
 \end{align*}

We can also bound each coefficient by 1. Indeed, rewriting the expression above using a matrix product and since the Frobenius norm is sub-multiplicative: 
$$a_{n,m} =  \left|\left| U_m^T V_n\right|\right|_F^2 \leq \left|\left| U_m\right|\right|_F^2 \left|\left| V_n\right|\right|_F^2 \leq 1.  $$

Once again, this adjacency structure is invariant under multiplication of the weights by a positive scalar.

\subsection{Statistical Significance Threshold}

In order to distinguish signal from noise, we now wish to develop a probabilistic model for an adjacency coefficient $a=||X^T Y||^2$ under a null hypothesis of random uniform distribution of $X \in \R^c $ and ${Y} \in \R^{c\times K^2}$ on the unit spheres $\mathbb{S}^{c-1}$ and $\mathbb{S}^{(c\times K^2)-1}$ respectively\footnote{Notation: $\mathbb{S}^{(c\times K^2)-1} := \left\{ Y \in \R^{c\times K^2} | \sum_{i,j}{y_{i,j}^2} = 1 \right\}$}. When the number of channel $c$ is large enough, it becomes possible to approximate the distribution of $a$ by a rescaled $\chi_2$ distribution with $K^2$ degrees of freedom. 
\begin{theorem} 
\label{chi2_k2}If $X,Y\sim \mathcal{U}(\mathbb{S}^{c-1}) \otimes \mathcal{U}(\mathbb{S}^{(c\times K^2)-1})$ then we have the convergence in distribution:
$$ c ||X^TY||_F^2 
\underset{c \longrightarrow +\infty}{\Longrightarrow} 
\frac{1}{K^2} \chi_2(K^2). $$
\end{theorem}

\begin{remark} We will thus use the following approximation for adjacency coefficients: $$a:=||X^TY||_F^2 \sim \frac{1}{cK^2} \chi_2(K^2).  $$
\end{remark}

\begin{proof}
See Appendix, section \ref{proof_chi2_k2}
\end{proof}
Using this model, we can again compute a threshold above which coefficients are considered significant with a probability $p$ confidence. We provide computation of higher moments of $\langle X,Y \rangle$ in Appendix \ref{A:momentsconv}.

\subsection{Remarks}

\paragraph{The generalization of the SVR to convolutional layers induces an asymmetry between the input and the output}
This is related to the choice of the dimensions merged during the flattening operation. We could have instead chosen to view the convolution layer as $i$ maps from $F$ to $F^o$, which leads to merging the output dimension with the kernel dimensions of $T$. The SVD we would have obtained corresponds to first performing a linear combination of inputs, scaling the result, and only then performing convolutions to obtain the output. This alternative technique, which we call \textit{co-SVR,} seems somewhat less insightful, as it prefers features relevant for the output of layers over those that are relevant for the input. Since the network also has a natural asymmetry as data flows from input to output, it makes sense to favor one direction. However, the co-SVR may be insightful to analyze gradient flows, as it exactly corresponds to the SVR seen by the gradient during backpropagation. 

\paragraph{Spatial pooling can be ignored} Spatial pooling (average, max etc.) between layers does not affect the size of the parameter space. Hence, we can simply ignore pooling in the SVR representation.

\paragraph{Link to the theory of tensor factorization} There are several ways to generalize SVD for higher dimensional matrices (\textit{ie} tensors). The process
we described here essentially flattens the tensor to two dimensions in order to perform the classical SVD. This is closely related to the Tucker decomposition for 3D-tensors, also known as higher order SVD (HOSVD) \citep{hitchcock,tucker}. See \cite{Rabanser2017IntroductionTT} for a modern introduction to tensor decompositions. The essential differences are that we only perform the flattening along one dimension and that we make use of both left and right singular vectors obtained with the classical SVD.

\begin{figure*}[h!]
\centering
\begin{subfigure}{.5\textwidth}
  \centering
  \includegraphics[width=0.9\linewidth]{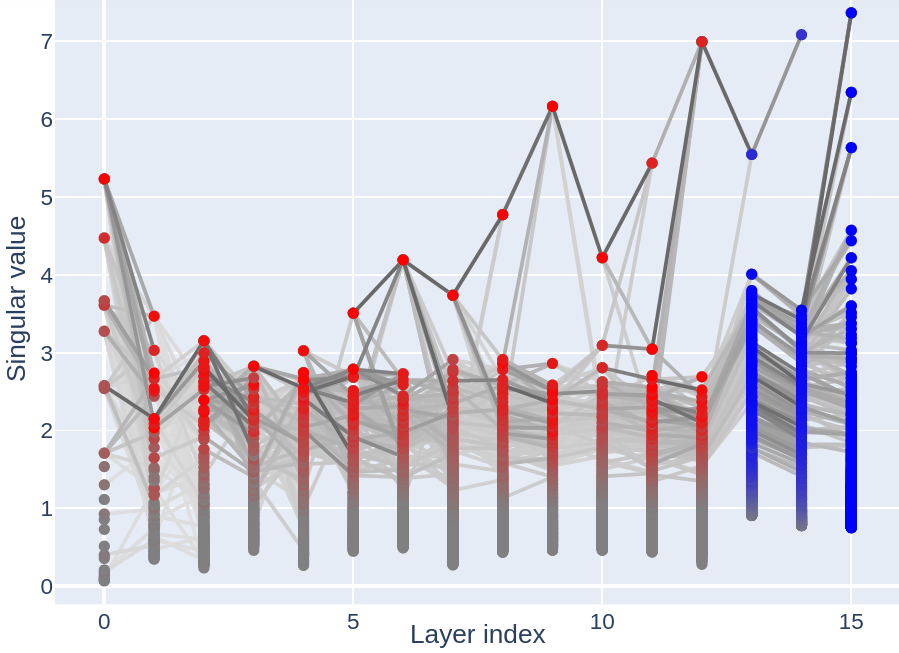}
  \caption{SVR of VGG16 - Red: Convolutional Layers, Blue: Fully Connected Layers}
  \label{vgg16}
\end{subfigure}%
\begin{subfigure}{.5\textwidth}
  \centering
  \includegraphics[width=0.9\linewidth]{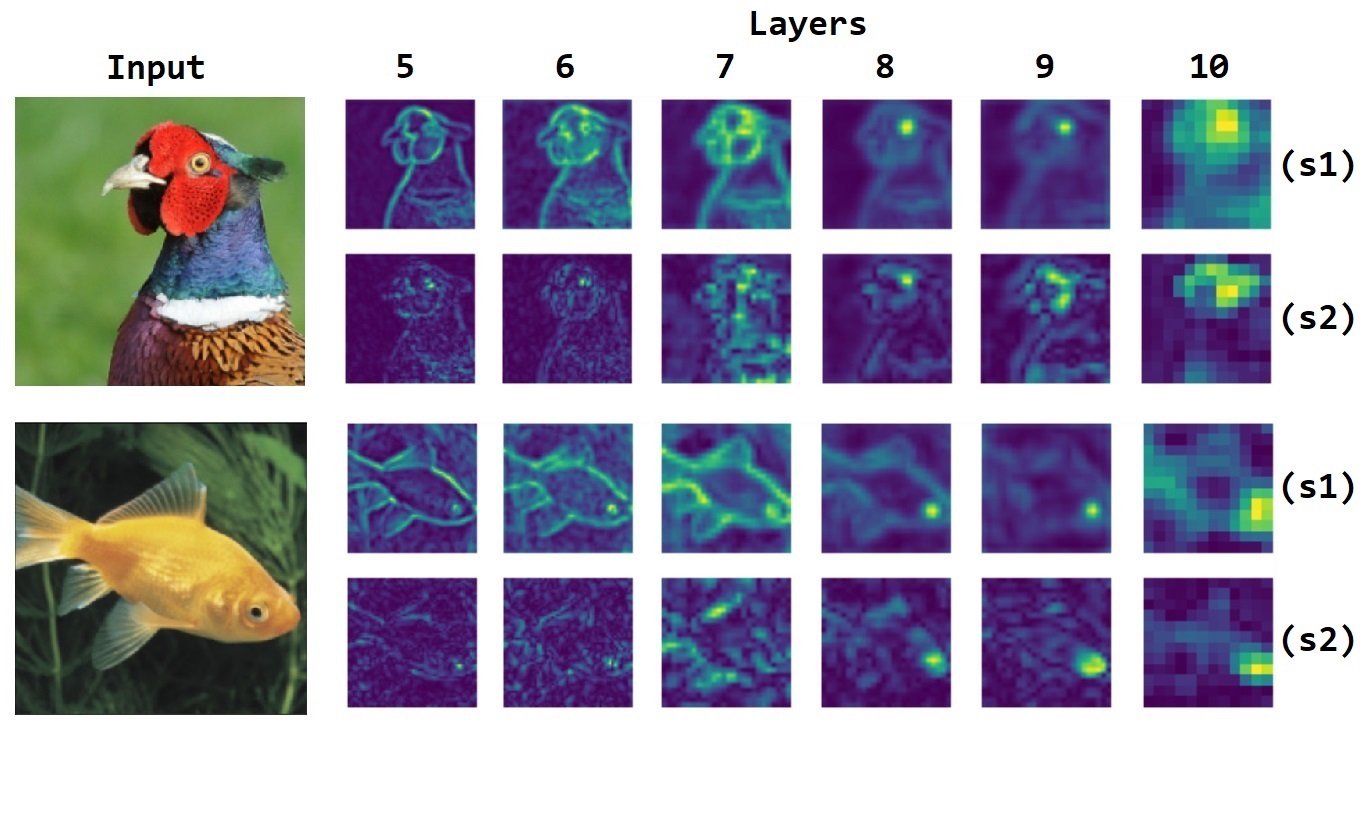}
  \caption{Activations of Spectral Neuron with Highest (s1) and Second Highest (s2) Singular Value}
  \label{coq_fish}
\end{subfigure}
\caption{The super-feature across layers in VGG networks highlights contours}
\label{vgg16superfeature}
\end{figure*}

\section{EXPERIMENTAL RESULTS}

As we explored SVR graphs, we swiftly encountered two phenomenons that caught our attention. We detail them in the next two sections. Those findings illustrate how the SVR can provide valuable insights into deep neural networks. We leave their detailed analysis for future work.

\subsection{The Super-Feature in VGG Networks}\label{sec:super-feature}

We computed the SVR for pretrained VGG networks on PyTorch 1.11.0 \citep{VGG,torch} which were trained on the ImageNet dataset \citep{ImageNet}. We noticed the presence of a "super-feature" in all VGG variants (11, 13, 16 and 19). It consists of spectral neurons with high singular values on consecutive layers that seem to connect to form a line. This is depicted in Figure \ref{vgg16}. To understand what this super-feature represents, recall that any spectral neuron in the SVR graph on a convolutional layer corresponds to a specific linear map $F^c \longrightarrow F$ where $F$ is a vector space of images of a given size and $c$ is the number of input channels. Thus, for a specific input, the activation on each spectral neuron corresponds to a single channel image. Visualizing these images can provide complementary insights to already existing visualization techniques \cite{voss2021visualizing}, on what information a specific spectral neuron is picking up. Figure \ref{coq_fish} depicts spectral activations obtained between layers 5 to 10 in VGG16 for two different inputs. Deeper layers are not depicted, because the size of the image become too small to be interpreted. The activation of spectral neurons of highest singular value (forming the super-feature) are compared with activations of the spectral neurons directly below. We clearly observe that the super-feature highlights the contours of the central object. 

We validated this qualitative observation by computing the cosine similarity between spectral images and a simple baseline for edge detection. More precisely, for a subsample of a 1000 pictures (one per class) from ImageNet \cite{ImageNetSample}, we plotted the average cosine similarity between spectral images (top-100 singular neurons) and the Sobel edge detector \footnote{For a given input $\mathbf{X}$, the Sobel edge detector is defined as: $ \sqrt{ \left( {\begin{bmatrix}-1&0&1\\-2&0&2\\-1&0&1\end{bmatrix}} \star {\mathbf  {X}} \right)^2 \quad + \left( {\begin{bmatrix}-1&-2&-1\\0&0&0\\1&2&1\end{bmatrix}} \star {\mathbf  {X}} \right)^2}. $} applied to input images\footnote{In order to match the dimensions of spectral images, we greyscaled and downsampled the input image through average pooling before applying the Sobel edge detector}.
We observe that the $0$-th spectral image is significantly more similar to the Sobel filter applied to the input image, than any of the other spectral images. This is depicted in figure \ref{fig:sobelcomparison} for Layer $6$, but the observation generalizes to other layers (see Appendix \ref{sec:edgedetector}).

\begin{figure}

  \includegraphics[width=\linewidth]{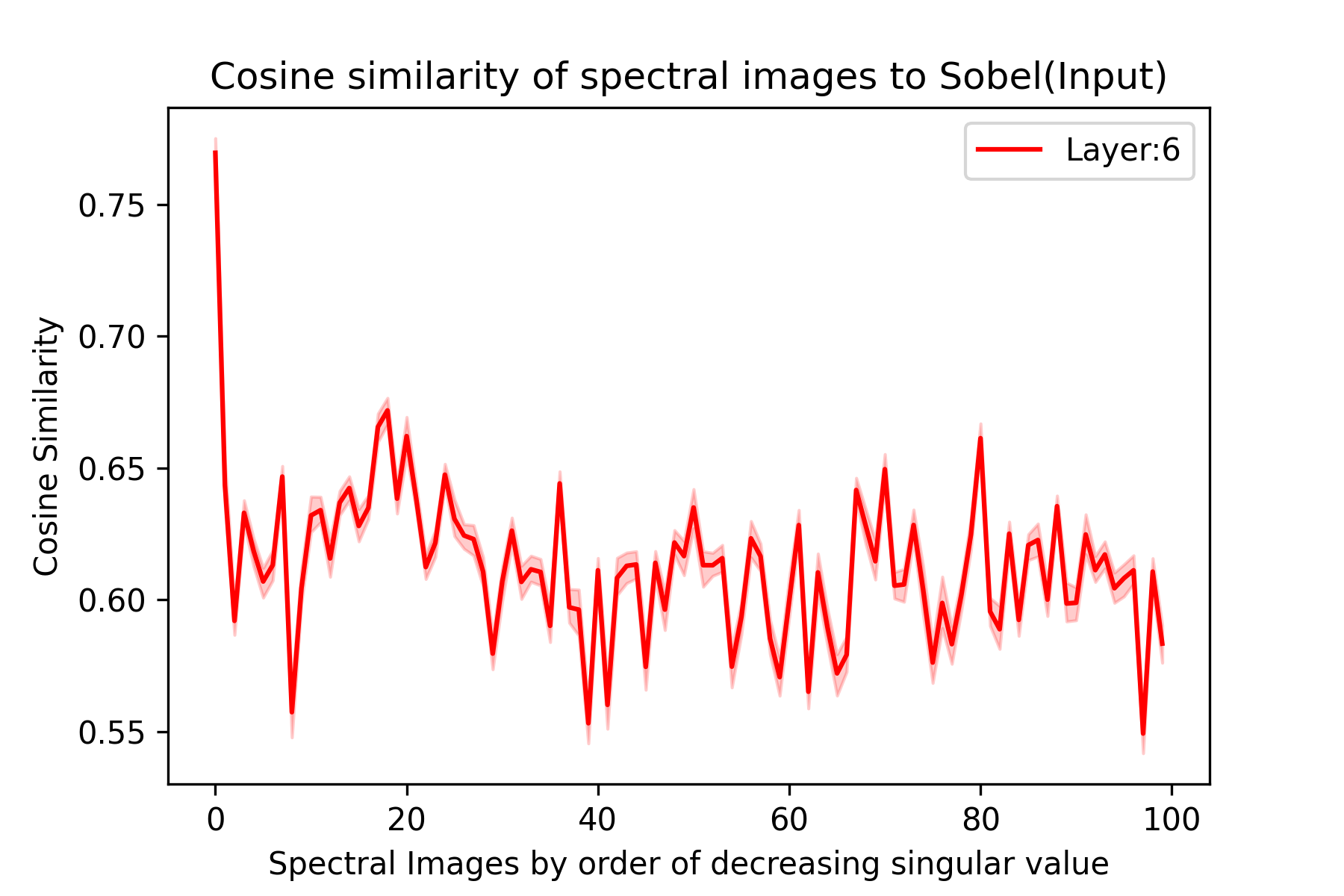}
  
\caption{Average (over 1000 images) Cosine similarity between Spectral Images (ordered by decreasing singular value), and Sobel filter of the input image. Red Shadow: 4 standard deviation interval.   }
\label{fig:sobelcomparison}
\end{figure}

\subsection{Deep Effect of Batch Normalization}

When computing the SVR of VGG networks that were trained using Batch Normalization \citep{batchnorm} and weight decay, we noticed a peculiar phenomenon. We could repeatedly observe significant connections between spectral neurons of \textbf{lowest singular value} (See Appendix, Figure \ref{vgg19_bn}). This was completely unexpected from the experiments we had done so far.
\subsubsection{Kernel Connections}

\begin{figure}
  \centering
  \includegraphics[width=0.66\linewidth]{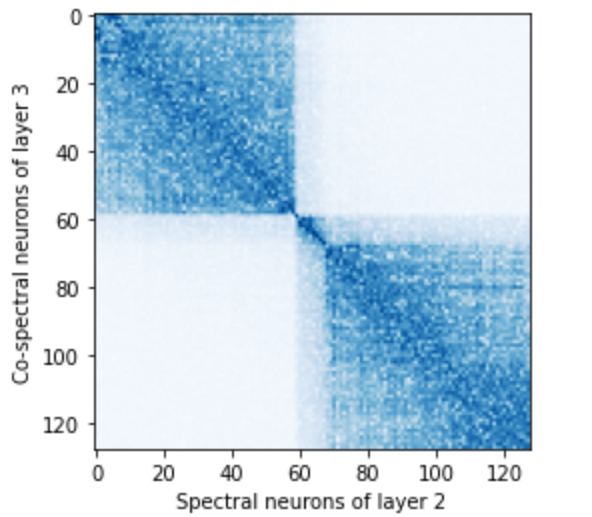}
  \caption{$|\widetilde{V}_{3}^T\bar{U}_2|$ - VGG19 Trained With Batch Norm}
  \label{adja_hosvd}
\end{figure}

 We observed this phenomenon both in the SVR and the co-SVR. This led us to construct layer-wise adjacency matrices that connect the SVR to the co-SVR. This new (symmetrical) construction turned out to capture the phenomenon even more vividly. 
More formally, given a weight tensor for a convolutional layer $T\in \R^{o \times i \times K \times K}$, the SVR flattens the tensor down to $\bar{T} \in \R^{o \times iK^2} $ while the co-SVR flattens it to $\widetilde{T}\in \R^{ o K^2 \times i}$. This then provides two SVD decompositions: $\bar{U}\bar{S}\bar{V}^T = \bar{T}$ and $\widetilde{U}\widetilde{S}\widetilde{V}^T = \widetilde{T}$. 
For two consecutive layers with weights $T_i$ and $T_{i+1}$, the matrix $\widetilde{V}_{i+1}^T\bar{U_i}$ represents the linear application that sends the activation of spectral neuron of layer $i$ to the activation of co-spectral neuron in layer $i+1$. Figure \ref{adja_hosvd} represents the absolute value of this matrix for layers 2 and 3 in the PyTorch pretrained version of VGG19 trained with batch normalization.\footnote{Those layers were chosen because the matrices associated were representative of the general phenomenon while being small enough to be effectively visualized.} The result is remarkably structured. One can observe a block-diagonal structure with 3 blocks. The two lowest blocks correspond to negligible (but measurably non-zero) singular values compared with the singular values associated to the indices of the first block. This plot highlights a remarkable correspondence between the near-kernels of successive layers, even though those spaces do not affect the output of the network by definition.

\subsubsection{Spontaneous Sparsification} 
\begin{figure}
  \centering
  \includegraphics[width=0.65\linewidth]{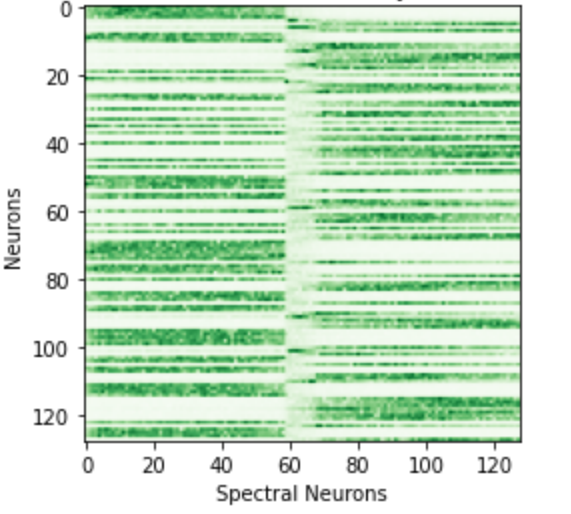}
  \caption{$|\bar{U}_2|$ - VGG19 Trained With Batch Norm}
  \label{U2}
\end{figure}

Remarkably, this phenomenon is also associated with a spectacular spontaneous sparsification of the weights. Figure \ref{U2} depicts the output representation of spectral neurons in the second layer ($\bar{U}_2$). One can see that each row, and thus each \textbf{usual} neuron, can be unambiguously mapped to one of the three blocks (corresponding to three groups of spectral neurons). Hence, all the neurons that are mapped to either the second or the third block can be pruned without changing the network behavior, since their output will be zeroed by negligible singular values. 
This spontaneous sparsification phenomenon was indirectly noticed by \citet{sparsity}. They identified the neurons to prune by selecting those for which activations over a large sample of data remained below a certain threshold. Our method, on the other hand, \textbf{does not rely on having any input data}. Surprisingly, their experiments seem to indicate that spontaneous sparsification does not occur when using SGD (unless the regularization is too strong, thus degrading accuracy). However, according to PyTorch reference training scripts (\href{https://github.com/pytorch/vision/tree/main/references/classification}{link here}), the model we used (\textit{vgg19\_bn}) was trained with default parameters which implies the use of the SGD optimizer. This suggests that this batch norm spontaneous sparsification could be a more general phenomenon than what was originally envisioned.  

Moreover, the SVR may be the right tool to understand the dynamic of this sparsification effect. We hypothesize that irrelevant features for the network are first connected together before being zeroed out by singular values and/or batch norm coefficients. 
The block structure in the near-kernel of the weights would thus be a trace of past dimension reductions performed by the network. 

This may provide useful research directions to understand the effectiveness of batch normalization for training deep neural networks.

\section{CONCLUSION}

The SVR offers a new perspective on representing the internal state of neural networks, by performing local SVD factorizations in each layers. Although it omits the non-linearities of the network, it allowed us to identify important properties of the learning process, for fully connected and convolutional networks of any size. This technique opens up new possibilities for machine learning research, by providing a rigorously defined summary of the learned weights structure in order to witness global phenomenons occurring across deep layers. 

We believe that the SVR could be a helpful tool in architecture design and hyperparameter optimization, as it can provide a more insightful overview of the internal state of a neural network. This would allow going beyond simple metrics based on accuracy and help to identify oversized layers, and may open up new possibilities to determine optimal \textbf{global} SVD compression schemes, which has been far less studied than layer-wise compression as noted by \citet{global_svd}. 

The SVR can also be used as a preprocessing tool of trained neural networks, in order to apply graph or topological techniques \cite{CTDA,BP22}  --that do not work at the usual scale of classical neural networks-- to analyze their weight structure. Indeed, thanks to the statistical framework we have developed, the SVR of a neural network usually has several orders of magnitude less edges than the initial network thanks to thresholding.

\subsubsection*{Acknowledgements}
The authors would like to thank Kathryn Hess Bellwald for her support on this project. 
This work received financial support from the Swiss Innovation Agency (Innosuisse project 41665.1 IP-ICT) and École Polytechnique (France).

\bibliography{biblio}


\appendix
\onecolumn

\section{MISSING PROOFS}

\subsection{Fourth Moment of $\langle X,Y \rangle$} \label{A:momentsFC}

We assume that $X,Y \sim \mathcal{U}(\mathbb{S}^{n-1}) \otimes \mathcal{U}(\mathbb{S}^{n-1}) $. We have already shown above that $\E(\langle X,Y\rangle^2) = \frac{1}{n}$. Moreover, because of sign symmetry, the third moment will be zero.
\begin{proposition}
\label{prop:mu4}We have: 
$$\mu_4 := \E(\langle X,Y\rangle^4) =  \frac{3}{n^2+2n}.$$ \end{proposition}

\begin{proof} Let us denote by $ \Gamma$ the covariance between two squared coordinates of $X$: $\Gamma := \E(x_1^2x_2^2)$. The first important remark is that because of permutation symmetry between the coordinates we have: 
$$\left\{
    \begin{array}{ll}
         \forall i \in \llbracket1,n\rrbracket, &\E(x_i^4)  =\mu_4  \\
         \forall i,j \in \llbracket1,n\rrbracket^2, &\  i \neq j \implies \E(x_i^2x_j^2) = \Gamma
    \end{array} 
    \right.
     $$
     
Because of rotational symmetry, we can replace $Y$ by any unitary vector without changing the distribution of $\langle X, Y \rangle$. Choosing $Y:= \frac{1}{\sqrt{n}}(1,1,...,1)$ yields an equation linking $\mu_4$ and $\Gamma$: 

\begin{equation}
\begin{split}
\mu_4 & = \E\left(\left(\frac{1}{\sqrt{n}} \sum x_i \right)^4\right)  \\
 & = \frac{1}{n^2} \left( \sum_{i} \E(x_i^4) + 3\E\left(\sum_{i \neq j}x_i^2x_j^2\right)   \right) \\ 
\mu_4 & = \frac{1}{n}\mu_4 + 3 \frac{n-1}{n}\Gamma.
\end{split}
\end{equation}
 The odd power terms in the expansion of the fourth power vanish because they have 0 expected value. 
 We can get a second equation linking $\mu_4$ and $\Gamma$ by leveraging the fact that $X$ is on the unit sphere: 
 \begin{equation} 
\begin{split}
1 & = \E\left[\left( \sum_{i} x_i^2\right)^2\right] \\
& = \E\left(\sum_{i} x_i^4 \right) + \E\left( \sum_{i\neq j} x_i^2 x_j^2 \right) \\ 
1 & = n\mu_4 + n(n-1)\Gamma.
\end{split}
\end{equation}
 
 This last equation gives: 
 $$\Gamma = \frac{1}{n-1} \left(\frac{1}{n}-\mu_4\right).$$
Injecting the expression for $\Gamma$ in the first equation we get:

$$\mu_4 = \frac{3}{n^2+2n}, \ \Gamma = \frac{1}{n^2+2n}. $$

\end{proof}

\begin{remark} The term $2n$ in the denominator can be viewed as a correction from the model $ \frac{1}{n}\chi_2(1)$ which approximates $\langle X,Y\rangle^2$ and which has a fourth moment of $\frac{3}{n^2}$. 
\end{remark}

\subsection{Standard Deviation of $r_1$} \label{A:momentsconv}
Let  ${Y} \in \R^{c\times K^2}$ be a random variable uniformly sampled on the unit sphere $\mathbb{S}^{(c\times K^2)-1}$. 
First, note that the columns of $Y$ can be rewritten: $\forall z \in \llbracket1,K^2\rrbracket, \ Y_z = r_z Q_z$ where $(Q_z) \in \R^c$ are independent and identically distributed according to $\mathcal{U}(\mathbb{S}^{o-1})$ and $r_z \in \R^+$ are identically distributed random variables such that $\sum_{z=1}^{K^2} r_z^2 = 1$. This formulates the idea that columns of $Y$ only interact through the amount of norm they consume. This statement can be easily proved by rewriting $Y$ as the normalization of a sample from the multivariate normal distribution. Using the linearity of the expected value we get that $\sum_{z=1}^{K^2} \E(r_z^2) = 1$ and hence $\E(r_1^2) = \frac{1}{K^2}$. We now wish to compute $\text{std}(r_1^2)$. We have the following proposition: 

\begin{proposition}
\label{prop:stdr1}
We have: $$ \text{std}(r_1^2) = \frac{1}{K^2}\sqrt{\frac{2K^2-2}{cK^2+2}} \simeq \sqrt{\frac{2}{c}}\E(r_1^2).$$
\end{proposition}
\begin{proof}
We start by computing the second moment of $r_1^2$: 
\begin{equation*}
\begin{split}
\E(r_1^4) & = \E\left( \left(\sum_{k=1}^c y_{k,1}^2\right)^2\right) \\
& = \E\left( \sum_{k=1}^c y_{k,1}^4\right)+\E\left( \sum_{k,l=1 ,k\neq l }^c y_{k,1}^2y_{l,1}^2\right) \\ 
& = c \mu_4 + c(c-1)\Gamma \\ 
& = \frac{c+2}{cK^4+2K^2}.
\end{split}
\end{equation*}
Where we used the expression for $\mu_4$ and $\Gamma$ derived in Proposition \ref{prop:mu4} with $n:=cK^2$. 
It follows that: 
\begin{equation*}
\begin{split}
\text{Var}(r_1^4) & = \frac{c+2}{cK^4+2K^2} - \frac{1}{K^4} \\ 
& = \frac{1}{K^4}\frac{2K^2-2}{cK^2+2} \\ 
& \simeq \frac{2}{cK^4} 
\end{split}.
\end{equation*}

We finally get the claimed result: $\text{std}(r_1^2)/\E(r_1^2) \simeq \sqrt{\frac{2}{c}} $. \end{proof}

\subsection{Proof of Theorem \ref{chi2_1}}

\newtheorem*{T1}{Theorem~\ref{chi2_1}}

\begin{T1}
If $X\sim\mathcal{U}(\mathbb{S}^{n-1})$ and $Y \in \mathbb{S}^{n-1}$ then:
$$ n \langle X,Y\rangle^2 \underset{n \longrightarrow +\infty}{\Longrightarrow} \chi_2(1).$$ 
\end{T1}

\begin{proof}
\label{proof_chi2_1}
Thanks to rotational invariance of the problem, we only consider the case of $Y:=(1,0,...,0)$. 
Let us denote by $Z$ a random variable sampled according to the normal $n$-dimensional distribution $\mathcal{N}(0,I_n)$. The distribution of $Z/||Z||$ is concentrated on the unit sphere and is invariant under rotations, thus $Z/||Z|| \sim \mathcal{U}(\mathbb{S}^{n-1})$ and therefore $Z/||Z||$ has the same distribution than $X$. This observation allows us to obtain the limit distribution of $(x_i)$: 

$$ \sqrt{n} \langle X,Y\rangle = \sqrt{n}x_1 \sim \frac{\sqrt{n}}{||Z||}z_1 =  \left( \frac{1}{n}\sum_{i=1}^n z_i^2 \right)^{-\frac{1}{2}} z_1.$$ 

The law of large numbers guarantees the convergence $ \frac{1}{n}\sum_{i=1}^n z_i^2 \underset{n \rightarrow +\infty}{\Longrightarrow} 1$ which implies, using Slutsky's theorem, that $\sqrt{n} x_1$ converges in distribution towards $\mathcal{N}(0,1)$ when $n$ is large. Since $\chi_2(1)$ is simply the distribution of a squared normal random variable, this proves the claimed result. 
\end{proof}

\subsection{Proof of Theorem \ref{chi2_k2}}

\newtheorem*{T2}{Theorem~\ref{chi2_k2}}

\begin{T2}
If $X\sim \mathcal{U}(\mathbb{S}^{c-1}) $ and ${Y\sim \mathcal{U}(\mathbb{S}^{(c\times K^2)-1})}$: 
$$ c ||X^TY||_F^2 
\underset{c \longrightarrow +\infty}{\overset{\mathbb{P}}{\Longrightarrow}} 
\frac{1}{K^2} \chi_2(K^2) $$.
\end{T2}

\begin{proof}
\textit{Some details are intentionally omitted to favor intuition and to make the proof lighter.} \\ 
First, just as in Proposition \ref{prop:stdr1}, let us rewrite columns of $Y$ in the form: $\forall z \in \llbracket1,K^2\rrbracket, \ Y_z = r_z Q_z$ where ${(Q_z) \in \R^c}$ are independent and identically distributed according to $\mathcal{U}(\mathbb{S}^{o-1})$ and $r_z \in \R^+$ are identically distributed random variables such that $\sum_{z=1}^{K^2} r_z^2 = 1$. Moreover we can show that $\text{std}(r_1^2)/\E(r_1^2)\simeq \sqrt{\frac{2}{c}}$ (see Proposition \ref{prop:stdr1}) which means that the standard deviation of $r_z$ becomes negligible compared to its expected value when the number of channel $c$ is large. We will thus use the expected value as an approximation: $$\forall z \in \llbracket1,K^2\rrbracket, \ r_z^2 \simeq \frac{1}{K^2}.$$ 
This approximation effectively removes the dependency between columns of $Y$. The adjacency coefficient $a$ thus becomes the sum of independent identically distributed random variable in the limit ${c\longrightarrow +\infty}$: 
$$ a =\sum_{1 \leq z \leq K^2} r_z^2 \langle X , Q_z \rangle^2 \simeq \frac{1}{K^2} \sum_{1 \leq z \leq K^2}  \langle X , Q_z \rangle^2.  $$ 
The independence comes from the the rotational invariance of the distribution of $Q_z$ which allows to replace $X$ in every term by any normal vector. From Theorem \ref{chi2_1}, we know that $\langle X, Q_z \rangle \sim \frac{1}{c} \chi_2(1)$. This implies that $a$ will follow a rescaled $\chi_2$ distribution with $K^2$ degrees of freedom: 

$$ a \sim \frac{1}{cK^2} \chi_2(K^2). $$
\label{proof_chi2_k2}
\end{proof}

\section{ADDITIONAL EXPERIMENTS}

\subsection{Internal Dimensions - Experimental Validation}
\label{sec:internal_dim_validation}

The definition of internal dimensions may seem slightly counter-intuitive at first. We provide here a small experiment to show that in a simple case, internal dimensions can recover the dimensions of a block structure in an orthogonal matrix.

\begin{figure}
  \includegraphics[scale=0.8]{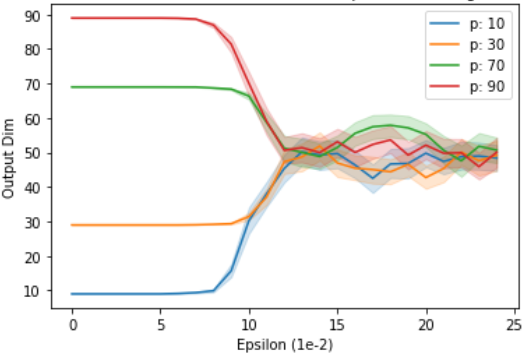}
\caption{Estimated Internal Output Dimension As Function Of $\epsilon$ ($n=100$) - Shadow : 95\% Confidence Interval}
\label{fig:internal_dim_val}
\end{figure}

Let us consider two integers $n \in \N^*$ and $p<n$. 
We sample independently two orthogonal matrices $Q_p$ and $Q_{n-p}$ according to the uniform distribution over $O_p(\R)$ and $O_{n-p}(\R)$. We consider the following block matrix : 
$$P:=\left (
   \begin{array}{c|c}
      Q_p & 0 \\
      \hline
      0 & Q_{n-p}\\ 
   \end{array}
\right). $$
$P$ is itself an orthogonal square matrix. In order to add noise to it, we use the Lie Algebra of $O_n(\R)$ which is the set of anti-symmetric matrices of size $n$. For $\epsilon >0$ we define the following random matrix: 
$$P(\epsilon) = P e^{\epsilon(A-A^T)}. $$

where $A \sim \mathcal{N}(0,I_n)$. $P(\epsilon)$ is also an orthogonal matrix which will keep a strong block structure as long as $\epsilon$ is small enough. We let $B(\epsilon)$ be the square matrix such that : 
$$B(\epsilon)_{i,j} := \sum_{k\leq i}\sum_{l \leq j} \left(P(\epsilon)_{k,l}\right)^2.$$ 
In Figure \ref{fig:internal_dim_val}, we plot the output dimension $d_{\text{out}}(\epsilon),\_ := \underset{i,j}{\text{argmax}} \left( B_{i,j}-\frac{1}{n}ij \right)$ as a function of $\epsilon$ for different values of $p$ (and $n=100$). The problem is completely symmetric for the input dimension. 
We observe that the internal output dimension does indeed recover the dimension of the first block up to a certain noise threshold above which it fluctuates around a value of $n/2$. 

\subsection{Survival Function of Adjacency Coefficients}

\begin{figure}[h]
\centering
\captionsetup[subfigure]{justification=centering}
\begin{subfigure}{.4\textwidth}

  \centering
  \includegraphics[width=1\linewidth]{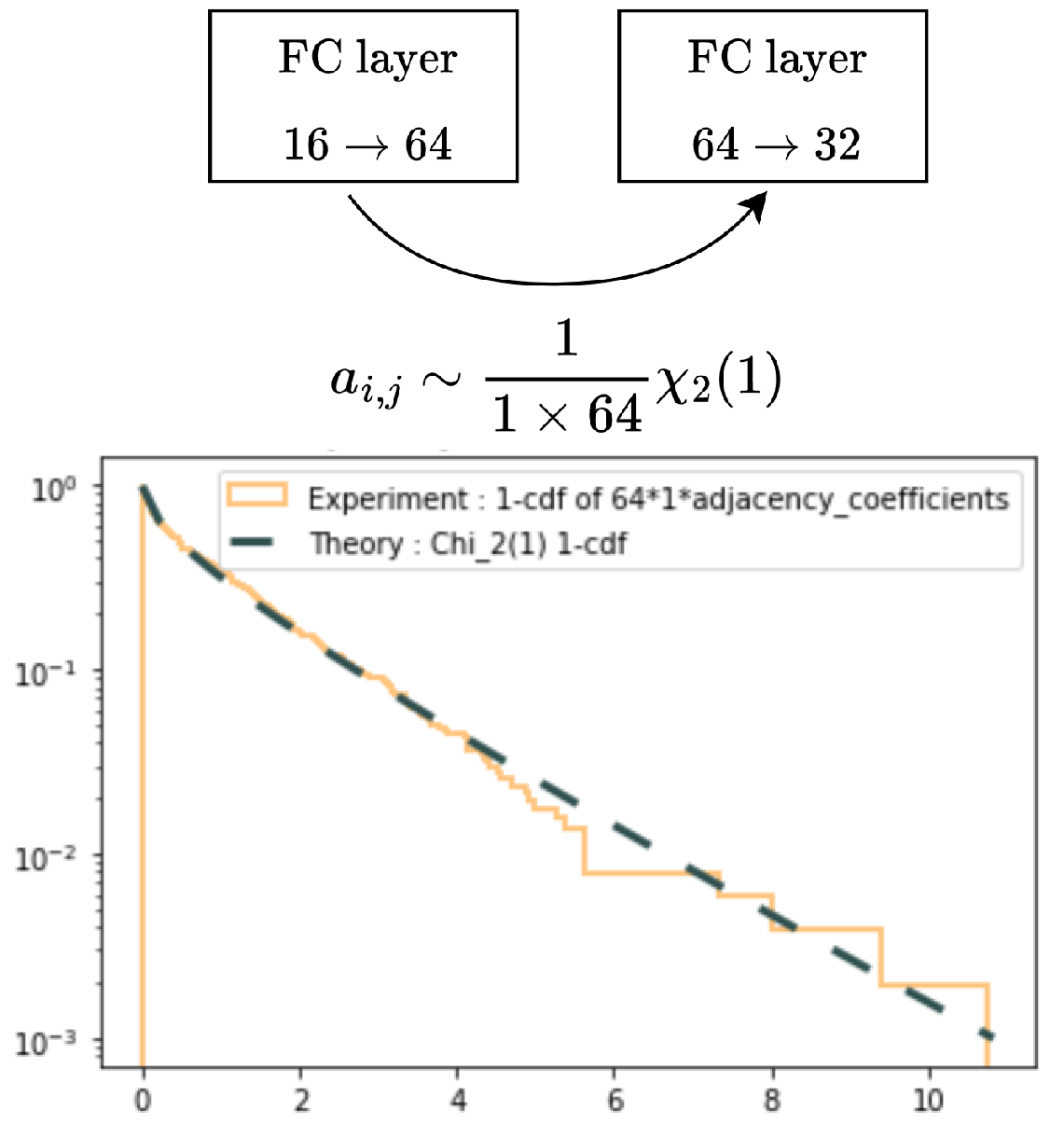}
  \caption{Adjacency coefficients between two fully connected layers: $\text{FC}(16,64)\rightarrow \text{FC}(64,32)$}
  \label{fig:sub1}
\end{subfigure}%
\begin{subfigure}{.4\textwidth}
  \centering
  \includegraphics[width=1\linewidth]{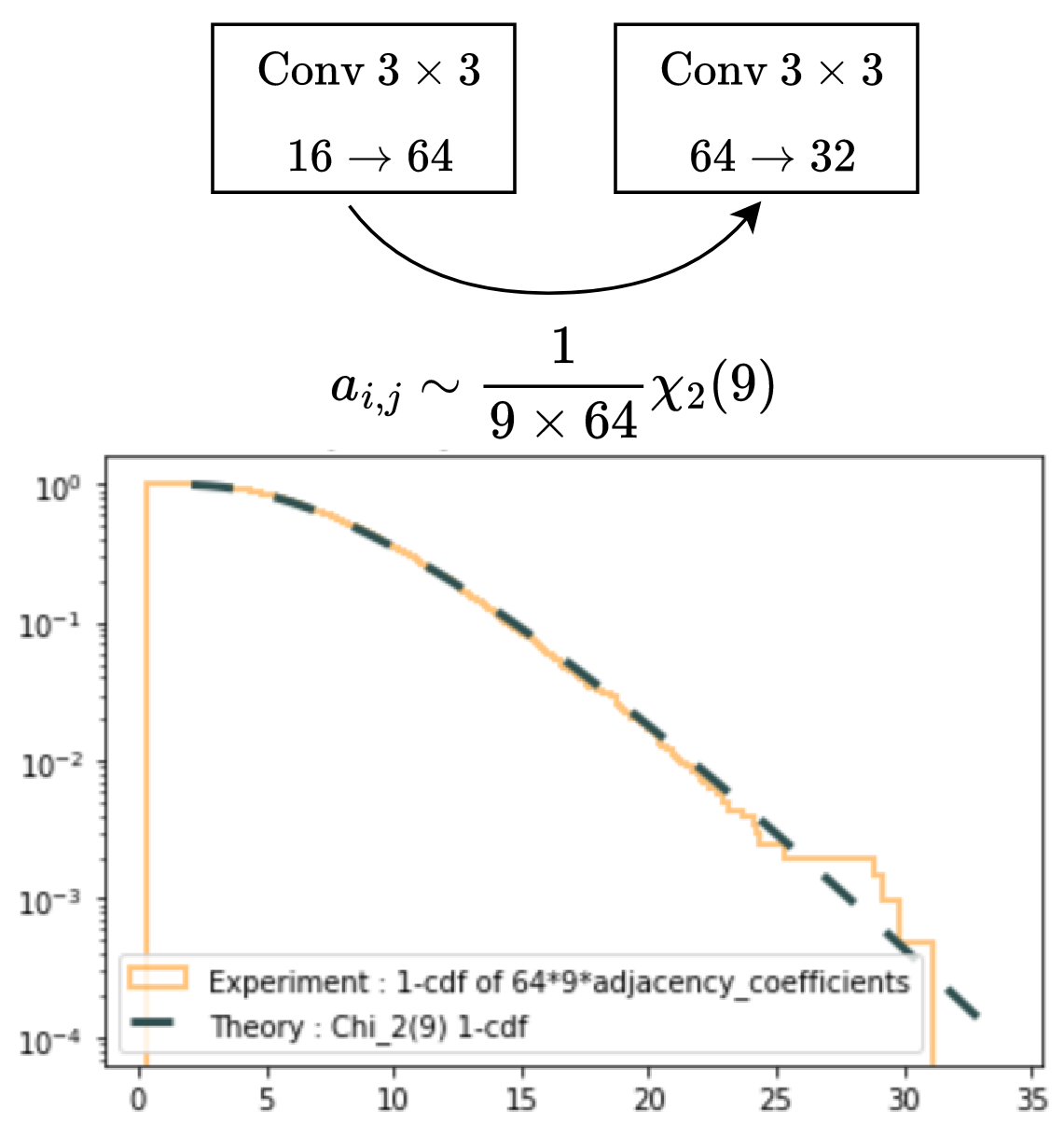}
  \caption{Adjacency coefficients between two convolutional layers: $\text{Conv}(16,64,3\times3)\rightarrow \text{Conv}(64,32,3\times3)$}
  \label{fig:sub2}
\end{subfigure}
\caption{Survival function $(1-\text{cdf}(x))$ of adjacency coefficients in different settings for a network with weights randomly sampled from the normal distribution}
\label{chi2_survival}
\end{figure}

\begin{figure}[h]
\centering
\captionsetup[subfigure]{justification=centering}
\begin{subfigure}{.4\textwidth}

  \centering
  \includegraphics[width=1\linewidth]{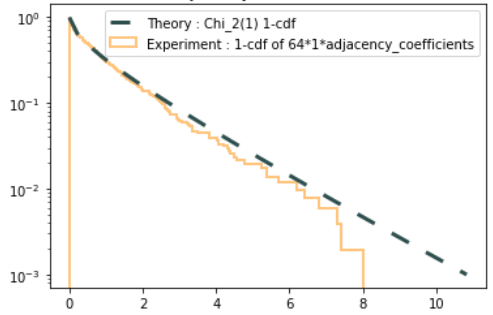}
  \caption{Adjacency coefficients between two fully connected layers: $\text{FC}(16,64)\rightarrow \text{FC}(64,32)$}
  \label{fig:sub1}
\end{subfigure}%
\begin{subfigure}{.4\textwidth}
  \centering
  \includegraphics[width=1\linewidth]{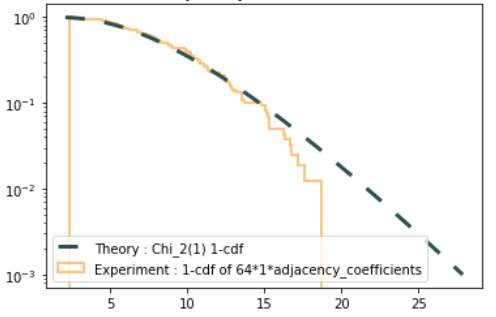}
  \caption{Adjacency coefficients between two convolutional layers: $\text{Conv}(16,64,3\times3)\rightarrow \text{Conv}(64,32,3\times3)$}
  \label{fig:sub2_uniform}
\end{subfigure}
\caption{Survival function $(1-\text{cdf}(x))$ of adjacency coefficients in different settings for a network with weights randomly sampled from the \textbf{uniform} distribution}
\label{chi2_survival_uniform}
\end{figure}

To validate the probabilistic models we developed for adjacency coefficients in a realistic setting, we performed the following experiment. Matrices of two layers (either convolutional or fully connected) were randomly initialized with independent coefficients sampled according to the normal distribution. The SVR was then computed to obtain the coefficients in the adjacency matrices. Since we aim to model the highest values yielded by this distribution, we compare how the survival function (one minus the cumulative distribution function) matches with the appropriately re-scaled $\chi_2$ distribution (Figure \ref{chi2_survival}).
Since the uniform distribution is also a common scheme for initializing weights, we performed the same experiments with a tensor whose weights were initialized according to a uniform distribution on $[-1,1]$ (Figure \ref{chi2_survival_uniform}). 

\captionsetup[figure]{justification=centering}

\captionsetup[figure]{justification=centering}

\subsection{SVR of VGG19 With and Without Batch Norm}

\begin{figure}[h!]
\centering
\captionsetup[subfigure]{justification=centering}
\begin{subfigure}{.5\textwidth}

  \centering
  \includegraphics[width=1\linewidth]{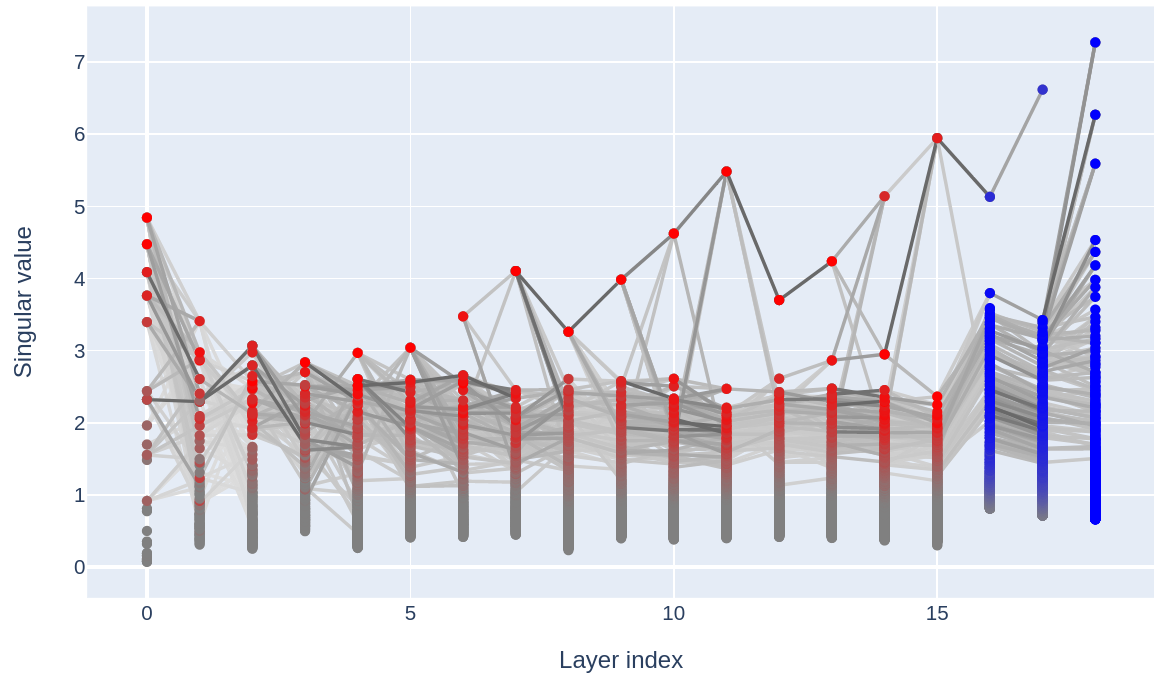}
  \caption{SVR of VGG19}
  \label{vgg19}
\end{subfigure}%
\begin{subfigure}{.5\textwidth}
  \centering
  \includegraphics[width=1\linewidth]{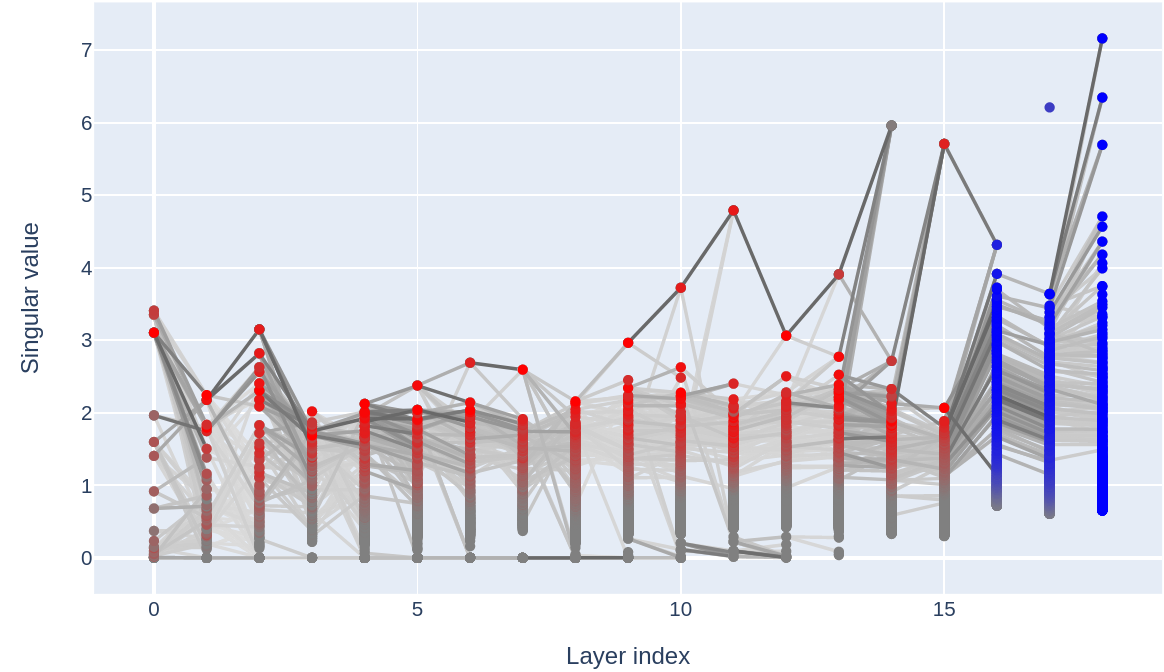}
  \caption{SVR of VGG19 trained with Batch Norm}
  \label{vgg19_bn}
\end{subfigure}
\caption{VGG19 With And Without Batch Normalization}
\label{vgg19_svr}
\end{figure}

\begin{figure}[h!]
\centering
\captionsetup[subfigure]{justification=centering}
\begin{subfigure}{.5\textwidth}

  \centering
  \includegraphics[width=1\linewidth]{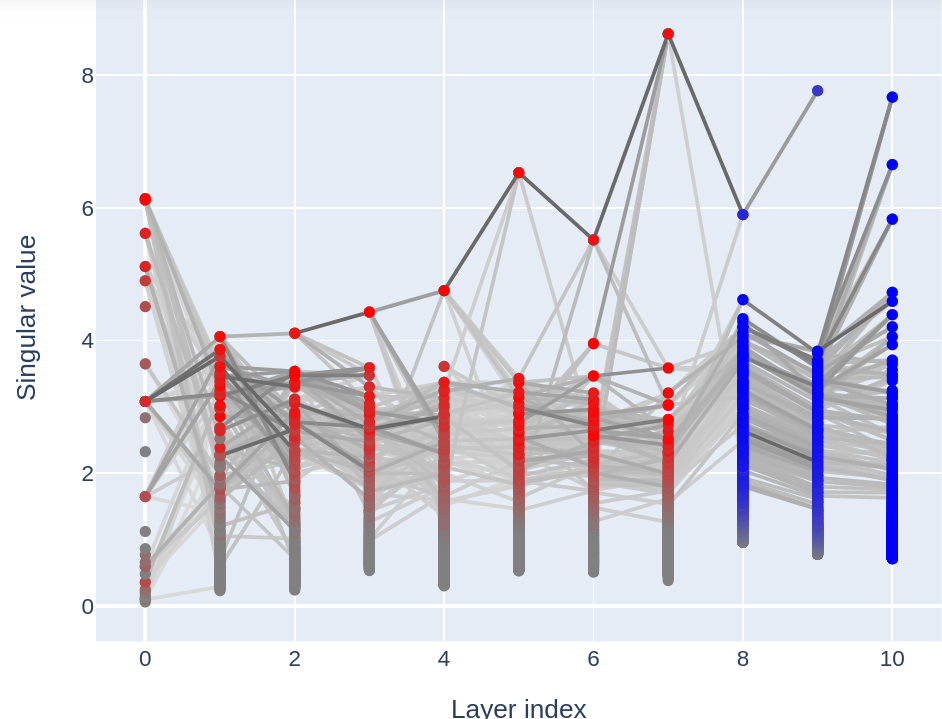}
  \caption{SVR of VGG11}
  \label{vgg11}
\end{subfigure}%
\begin{subfigure}{.5\textwidth}
  \centering
  \includegraphics[width=1\linewidth]{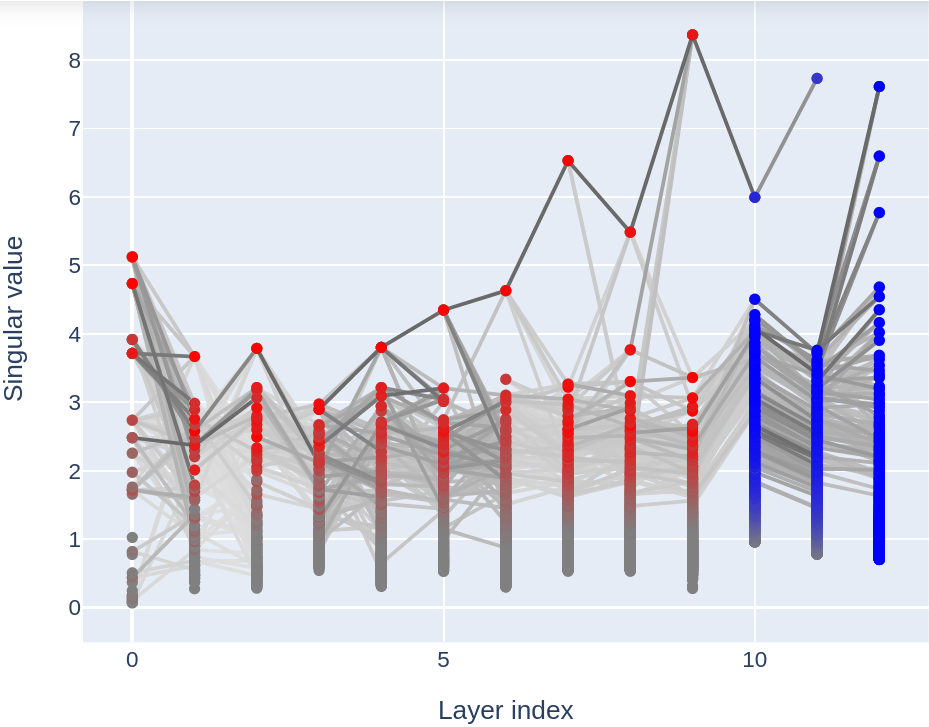}
  \caption{SVR of VGG13}
  \label{vgg13}
\end{subfigure}
\caption{Comparison between the SVR of VGG11 and VGG13}
\label{vgg11_13}
\end{figure}

We compute the SVR for the pretrained version of VGG networks availabe on PyTorch. 
In Figure \ref{vgg19_svr} we depict the SVR for both VGG19 (\ref{vgg19}) and VGG19 with batch normalization (\ref{vgg19_bn}). For the latter, we ignored batch norm scaling coefficients to allow for a more faithful comparison with the first case. We observe connections between the near-kernels of layers when batch norm is introduced. 
In Figure \ref{vgg11_13}, we compare VGG11 and VGG13 networks, in both cases the super-feature noticed in section \ref{sec:super-feature} can be observed.  

\captionsetup[figure]{justification=centering}

\captionsetup[figure]{justification=centering}

\subsection{The Super-Feature in VGG Networks Is An Edge Detector}
\label{sec:edgedetector}

\begin{figure}[h!]
\centering
\captionsetup[subfigure]{justification=centering}
\begin{subfigure}{.5\textwidth}

  \centering
  \includegraphics[width=1\linewidth]{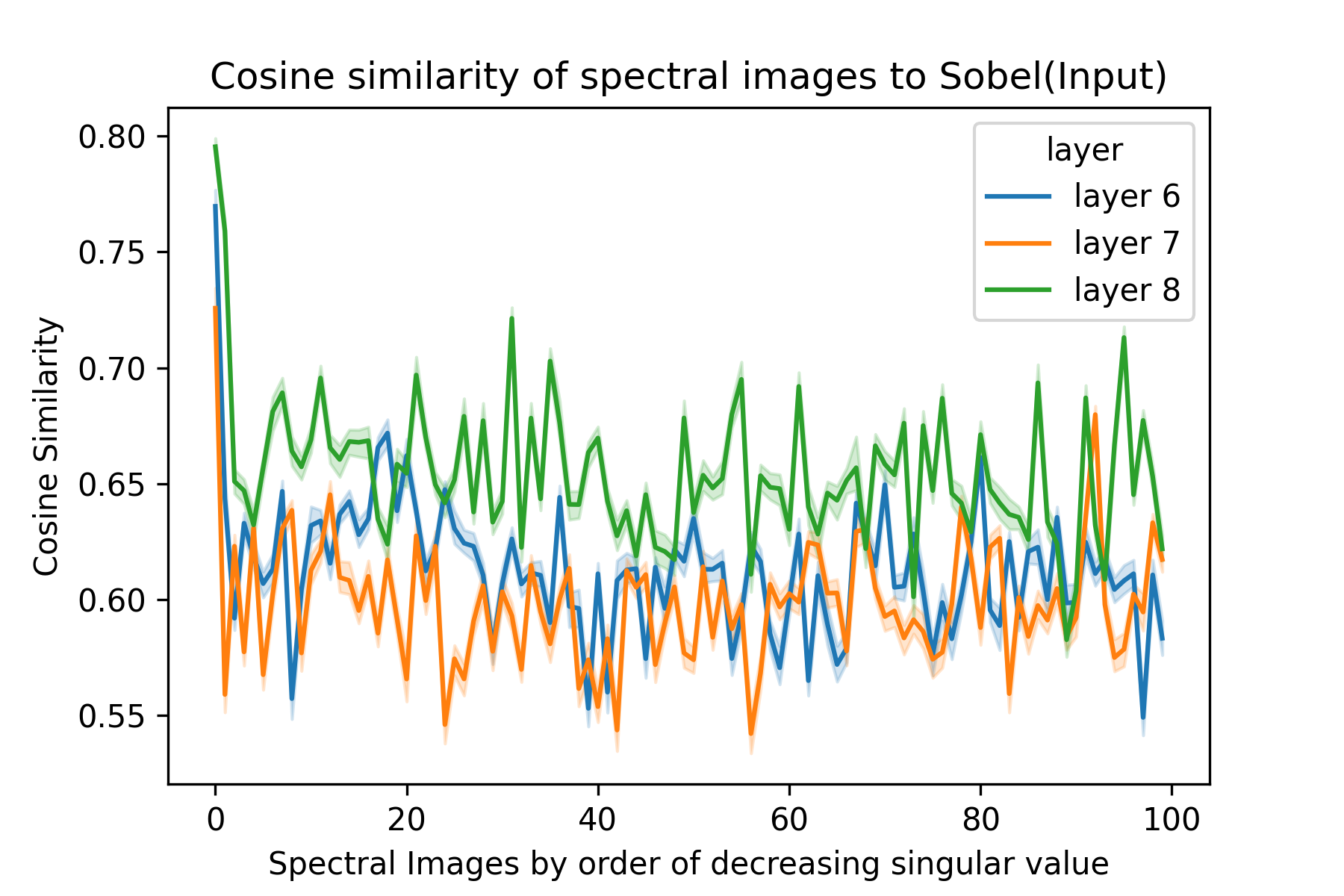}
  \caption{Layers 6,7 and 8}
\end{subfigure}%
\begin{subfigure}{.5\textwidth}
  \centering
  \includegraphics[width=1\linewidth]{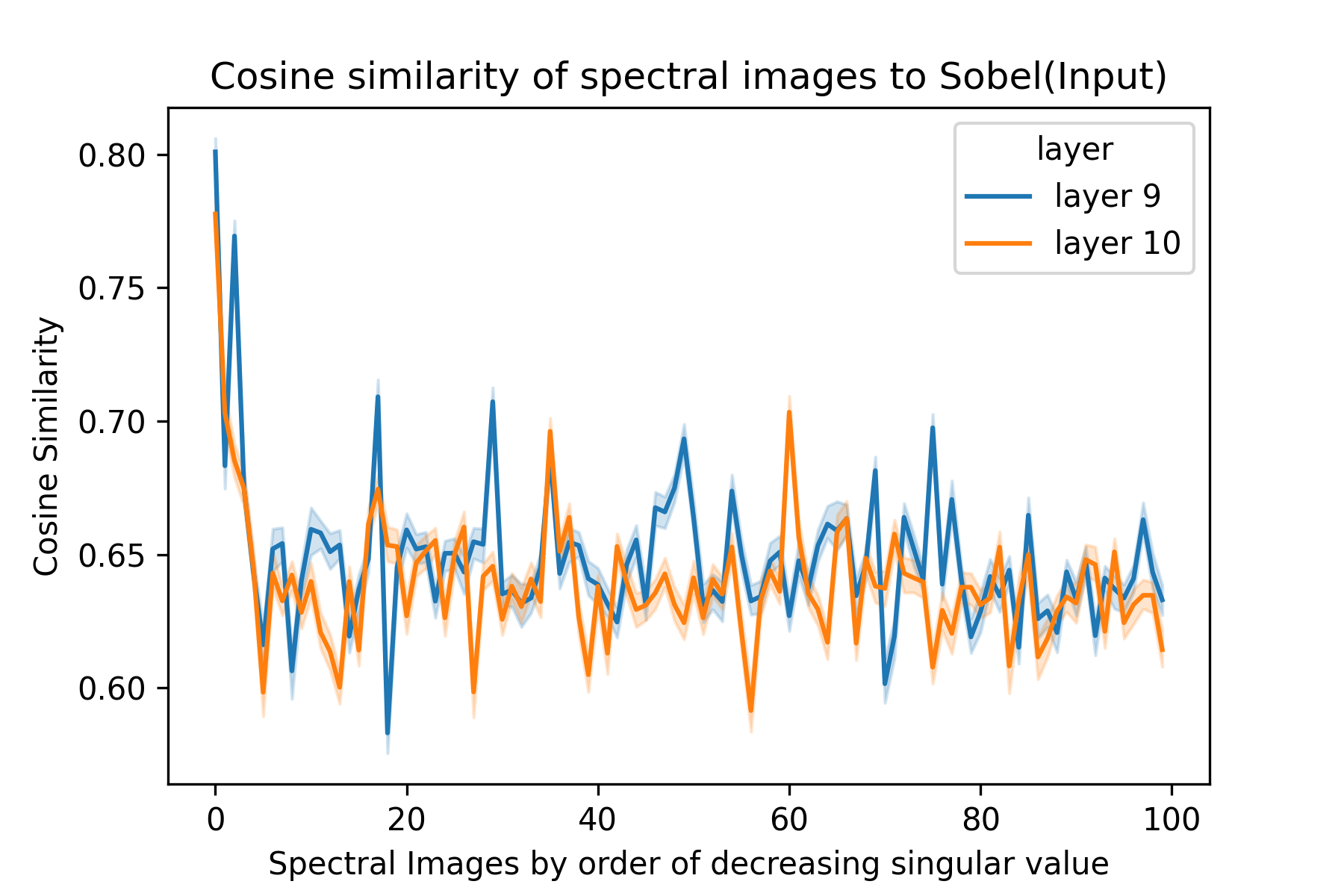}
  \caption{Layers 9 and 10}
\end{subfigure}
\caption{Average (Over 1000 Images) Cosine Similarity Between Spectral Images (ordered by decreasing singular value) And The Sobel Edge Detector Applied To The
Input Image. Shadow: 4 Standard Deviation Interval}
\label{fig:sobel}
\end{figure}

We provide quantified numerical evidence for the "super-feature" of VGG networks to be an edge detector. To do so, we downscaled the input through average pooling down to the dimension of spectral images of a given layer, we then transformed the image into grey scale and applied the Sobel filter to it\footnote{For a given input $\mathbf{X}$, the Sobel filter is defined as : $ \sqrt{ \left( {\begin{bmatrix}-1&0&1\\-2&0&2\\-1&0&1\end{bmatrix}} \star {\mathbf  {X}} \right)^2 \quad + \left( {\begin{bmatrix}-1&-2&-1\\0&0&0\\1&2&1\end{bmatrix}} \star {\mathbf  {X}} \right)^2} $.}. This constitutes a fair proxy for edge detection. We then  plotted the cosine similarity (averaged over 1000 images, one per class) between the absolute value of spectral images and this proxy in Figure \ref{fig:sobel}. We clearly observed that for all layers where the super-feature is present, the first spectral image reaches the highest cosine similarity.

\captionsetup[figure]{justification=centering}

\section{A COMPUTATIONAL SUMMARY OF THE SVR }

Table \ref{tab:fc-cnn} summarizes the main resemblances and differences between the SVR computation for fully connected and convolutional layers. 

\begin{table*}[h]
\begin{center}
\caption{Comparison Between Fully Connected And Convolutional Layers}
\label{tab:fc-cnn}
\begin{tabular}{ |c|c|c| } 
\hline
\textbf{Layer type} & \textbf{Fully connected} & \textbf{Convolutional} \\
\hline
\hline
&&\\
Input and output space & $\R^i \longrightarrow \R^o $ & $F^i \longrightarrow F^o $ \\ 
&&\\
\hline
&&\\
Factoring the output dimension & $o \times \left( \R^i \longrightarrow \R \right) $  & $o \times \left( F^i \longrightarrow F \right)$ \\  
\hline
&&\\
Fundamental operation & Scalar product $\cdot $& 2D-convolution $\star$ \\ 
\hline 
Building block & $\begin{array}{ccccc}
& & \R^i & \to & \R \\
 & & x & \mapsto & \theta \cdot x  \\
\end{array}$ & $\begin{array}{ccccc}
& & F^i & \to & F \\
 & & x & \mapsto & \theta \star x  \\
\end{array}$\\ 
\hline
Parameter space of one building block& &\\
$K$ \textit{denotes the kernel size (3 or 5 typically)}&$\theta \in \R^i $& $\theta \in \R^{i\times K \times K}$\\
\hline
 && \\ 
SVD && \\ \textit{The bar corresponds to the flattening operation} & $USV^T = \begin{pmatrix} \ &\theta_1^T &  \ \\ \  & ... & \  \\ \ & \theta_o^T \ & \end{pmatrix} $ & $US\bar{V}^T = \begin{pmatrix} \ &\bar{\theta_1}^T &  \ \\ \  & ... & \  \\ \ & \bar{\theta_o}^T \ & \end{pmatrix} $\\ 
\hline
&& \\ 
Recovering output $k \in \llbracket1,o\rrbracket $&  $\theta_k \cdot x = \sum\limits_{\substack{m}}  u_{k,m} s_m \left( V_m \cdot x  \right)$ & $\theta_k \star x = \sum\limits_{\substack{m}}  u_{k,m} s_m \left( V_m \star x  \right)$\\ 
\hline
Adjacency between spectral neuron $m$ and $n$ && \\ 
\textit{In this line, $V$ is from the next layer} & $ \left( U_m^T V_n\right)^2 $ & $\left|\left| U_m^T V_n\right|\right|_F^2 $ \\ 
\hline
Baseline distribution of adjacency coefficients &&\\
\textit{when $o\longrightarrow \infty$, with randomness assumptions}  & $\frac{1}{o} \chi_2(1)$ & $\frac{1}{oK^2} \chi_2(K^2)$\\ 

\hline 
\end{tabular}

\end{center}
\end{table*}
\end{document}